\documentclass[10pt,twocolumn,letterpaper]{article}

\usepackage{cvpr}
\usepackage{times}
\usepackage{graphicx}
\usepackage{amsmath}
\usepackage{amssymb}
\usepackage{float}
\usepackage{mathtools}
\usepackage{multirow}
\usepackage{booktabs}
\usepackage{subcaption}
\usepackage{colortbl, xcolor}
\definecolor{beforefail}{RGB}{106,160,232}
\definecolor{atfail}{RGB}{240,206,98}
\definecolor{afterfail}{RGB}{0, 140, 255}
\usepackage[LY1]{fontenc}
\pdfmapfile{+toonacious.map}
\usepackage{array}
\newcolumntype{P}[1]{>{\centering\arraybackslash}p{#1}}
\usepackage{xcolor}
\usepackage[hyphens]{url}
\usepackage[pagebackref=true,breaklinks=true,letterpaper=true,colorlinks,bookmarks=false]{hyperref}

\cvprfinalcopy 

\def\cvprPaperID{6398} 

\ifcvprfinal\fi
\begin{document}

\newcommand\oopsfont[1]{{\fontfamily{toonacious}\fontseries{m}\fontshape{n}\selectfont #1}}
\newcommand\figref[1]{Figure \ref{#1}}
\newcommand\oops{\oopsfont{\detokenize{_o_ops_!_  }}}

\title{\oops \ Predicting Unintentional Action in Video}
\author{Dave Epstein\hspace{2em}Boyuan Chen\hspace{2em}Carl Vondrick\\
Columbia University\\
\href{http://oops.cs.columbia.edu}{oops.cs.columbia.edu}
}

\maketitle

\enlargethispage{-7.2cm} 
\noindent\begin{picture}(0,0)
\put(0,-403){\begin{minipage}{\textwidth}
\centering
\includegraphics[width=\linewidth]{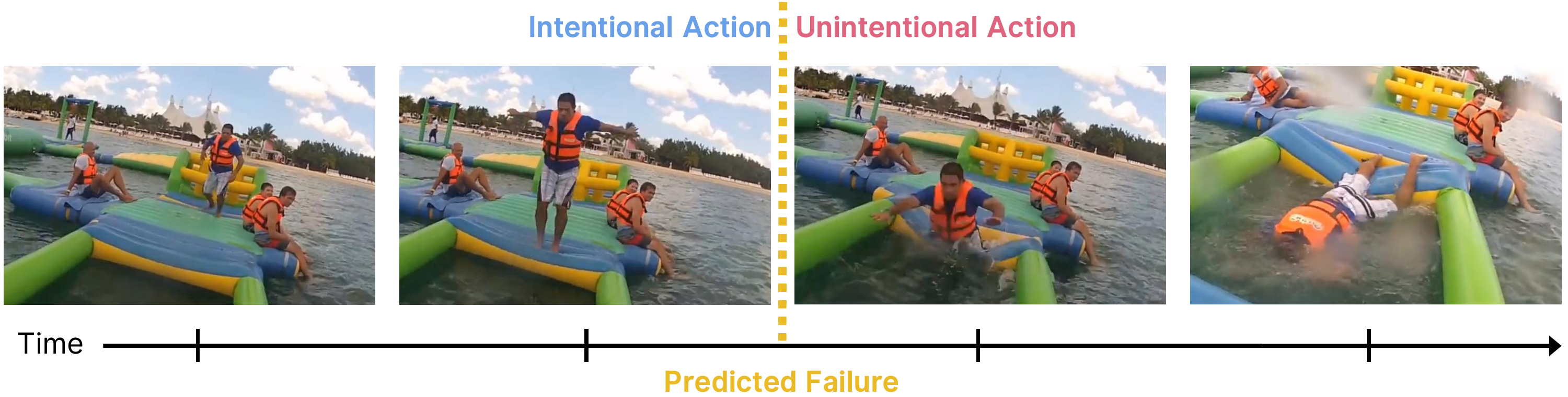}
\vspace{-2em}
\captionof{figure}{\textbf{Intentional versus Unintentional:} Did this person intend for this action to happen, or was it an accident? In this paper, we introduce a large in-the-wild video dataset of unintentional action. Our dataset, which we have collected by downloading ``fail'' videos from the web, contains over twenty thousand clips, and they span a diverse number of activities and scenes. Using this dataset, we study a variety of visual clues for learning to predict intentionality in video.}
\label{fig:teaser}
\end{minipage}}
\end{picture}%

\begin{abstract}

From just a short glance at a video, we can often tell whether a person's action is intentional or not. Can we train a model to recognize this? 
We introduce a dataset of in-the-wild videos of unintentional action, as well as a suite of tasks for recognizing, localizing, and anticipating its onset. We train a supervised neural network as a baseline and analyze its performance compared to human consistency on the tasks. We also investigate self-supervised representations that leverage natural signals in our dataset, and show the effectiveness of an approach that uses the intrinsic speed of video to perform competitively with highly-supervised pretraining. However, a significant gap between machine and human performance remains.




\end{abstract}


\vspace{-1em}

\section{Introduction}\label{sec:intro}

From just a glance at a video, we can often tell whether a person's action is intentional or not. For example, Figure \ref{fig:teaser} shows a person attempting to jump off a raft, but unintentionally tripping into the sea. In a classic series of papers, developmental psychologist Amanda Woodward demonstrated that this ability to recognize the intentionality of action is learned by children during their first year \cite{woodward1999infants,woodward2009infants,brandone2009you}. However, predicting the intention behind action has remained elusive for machine vision. 
Recent advances in action recognition have largely focused on predicting the physical motions and atomic actions in video \cite{kay2017kinetics,gu2018ava,monfort2019moments}, which captures the means of action but not the intent of action. 
\enlargethispage{-7.2cm} 

We believe a key limitation for perceiving visual intentionality has been the lack of realistic data with natural variation of intention. Although there are now extensive video datasets for action recognition \cite{kay2017kinetics,gu2018ava,monfort2019moments}, people are usually competent, which causes datasets to be biased towards successful outcomes. However, this bias for success makes discriminating and localizing visual intentionality difficult for both learning and quantitative evaluation. 

\begin{figure*}[t]
\includegraphics[width=\linewidth]{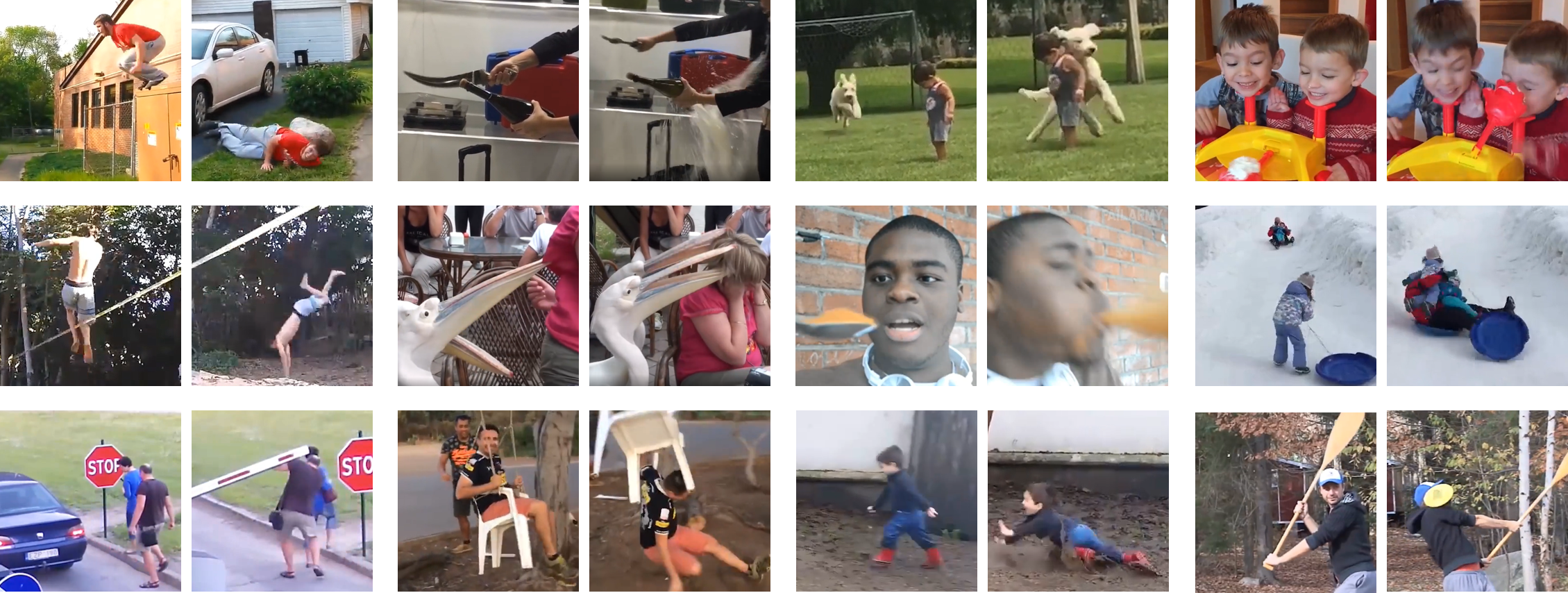}
\vspace{-2em}
\caption{\textbf{The \oops Dataset:}
Each pair of frames shows an example of intentional and unintentional action in our dataset. By crawling publicly available ``fail'' videos from the web, we can create a diverse and in-the-wild dataset of unintentional action. For example, the bottom-left corner shows a man failing to see a gate arm, and the top-right shows two children playing a competitive game where it is inevitable one person will fail to accomplish their goal.\vspace{-1em}}
\label{fig:dataset} 
\end{figure*}

We introduce a new annotated video dataset that is abundant with unintentional action, which we have collected by crawling publicly available ``fail'' videos from the web. Figure \ref{fig:dataset} shows some examples, which cover in-the-wild situations for both intentional and unintentional action.
Our video dataset, which we will publicly release, is both large (over 50 hours of video) and diverse (covering hundreds of scenes and activities). We annotate videos with the temporal location at which the video transitions from intentional to unintentional action. We define three tasks on this dataset: classifying the intentionality of action, localizing the transition from intentional to unintentional, and forecasting the onset of unintentional action shortly into the future. 

To tackle these problems, we investigate several visual clues for learning with minimal labels to recognize intentionality. First, we propose a novel self-supervised task to learn to predict the speed of video, which is incidental supervision available in all unlabeled video, for learning an action representation. Second, we explore the predictability of the temporal context as a clue to learn features, as unintentional action often deviates from expectation. Third, we study the order of events as a clue to recognize intentionality, since intentional action usually precedes unintentional action.

Experiments and visualizations suggest that unlabeled video has intrinsic perceptual clues to recognize intentionality. Our results show that, while each self-supervised task is useful, learning to predict the speed of video helps the most. By ablating model and design choices, our analysis also suggests that models do not rely solely on low-level motion clues to solve unintentional action prediction. Moreover, although human consistency on our dataset is high, there is still a large gap in performance between our models and human agreement, underscoring that analyzing human goals from videos remains a fundamental challenge in computer vision. We hope this dataset of unintentional and unconstrained action can provide a pragmatic benchmark of progress.

This paper makes two primary contributions. Firstly, we introduce a new dataset of unconstrained videos containing a substantial variation of intention and a set of tasks on this dataset. Secondly, we present models that leverage a variety of incidental clues in unlabeled video to recognize intentionality. The remainder of this paper will describe these contributions in detail. Section \ref{sec:related} first reviews related work in action recognition. Then, Section \ref{sec:dataset} introduces our dataset and summarizes its statistics. Section \ref{sec:learning} presents several self-supervised learning approaches to learn visual representations of intentionality. In Section \ref{sec:exp}, we present quantitative and qualitative experiments to analyze our model. We release all data, software, and models on the website.

\section{Related Work}\label{sec:related}

\begin{figure*}[t]
\begin{subfigure}[t]{0.33\linewidth}
\vskip 0pt
\includegraphics[width=\linewidth]{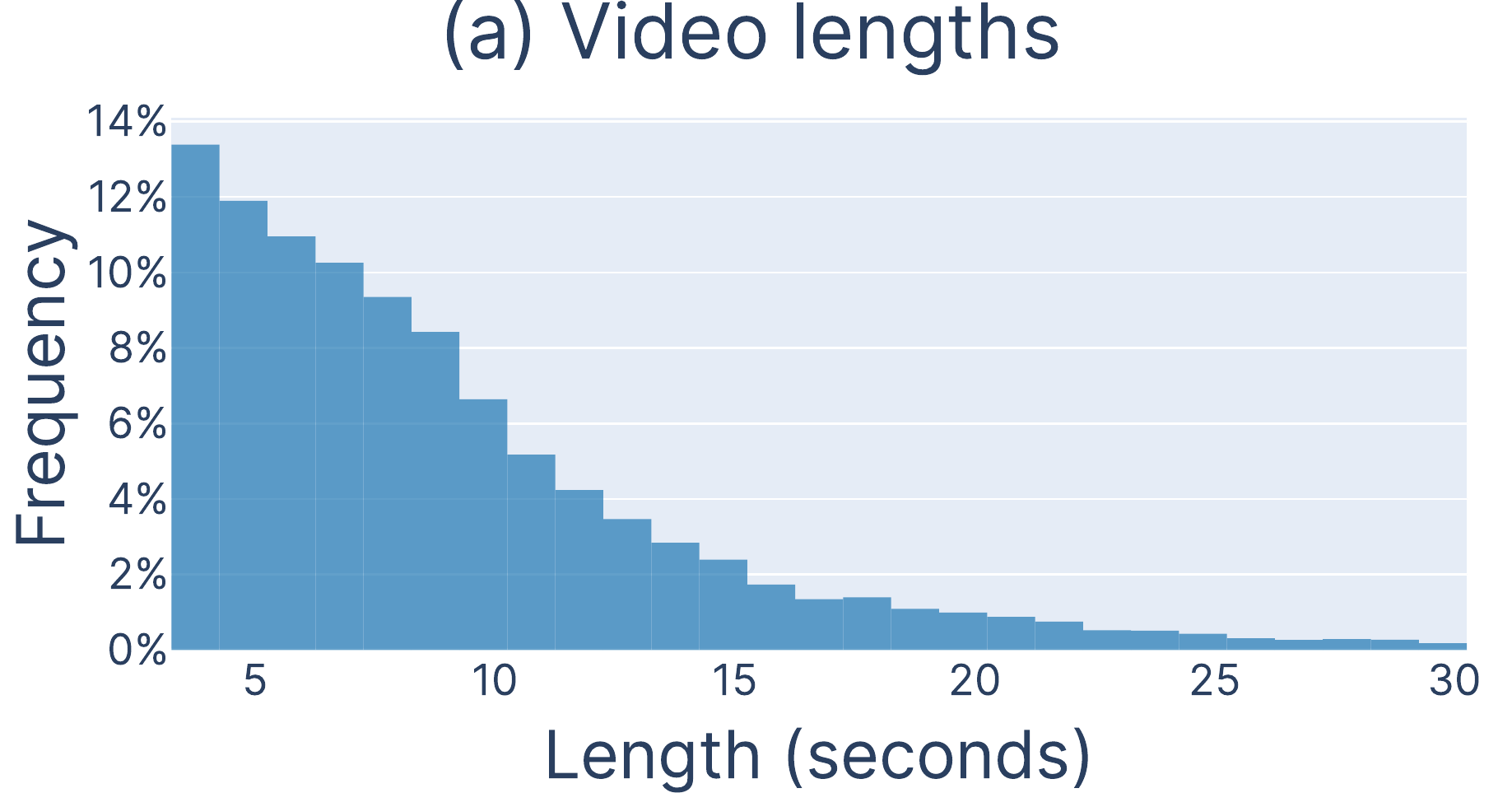}
\phantomsubcaption\label{fig:vid_len}
\end{subfigure}
\begin{subfigure}[t]{0.33\linewidth}\vskip 0pt
\includegraphics[width=\linewidth]{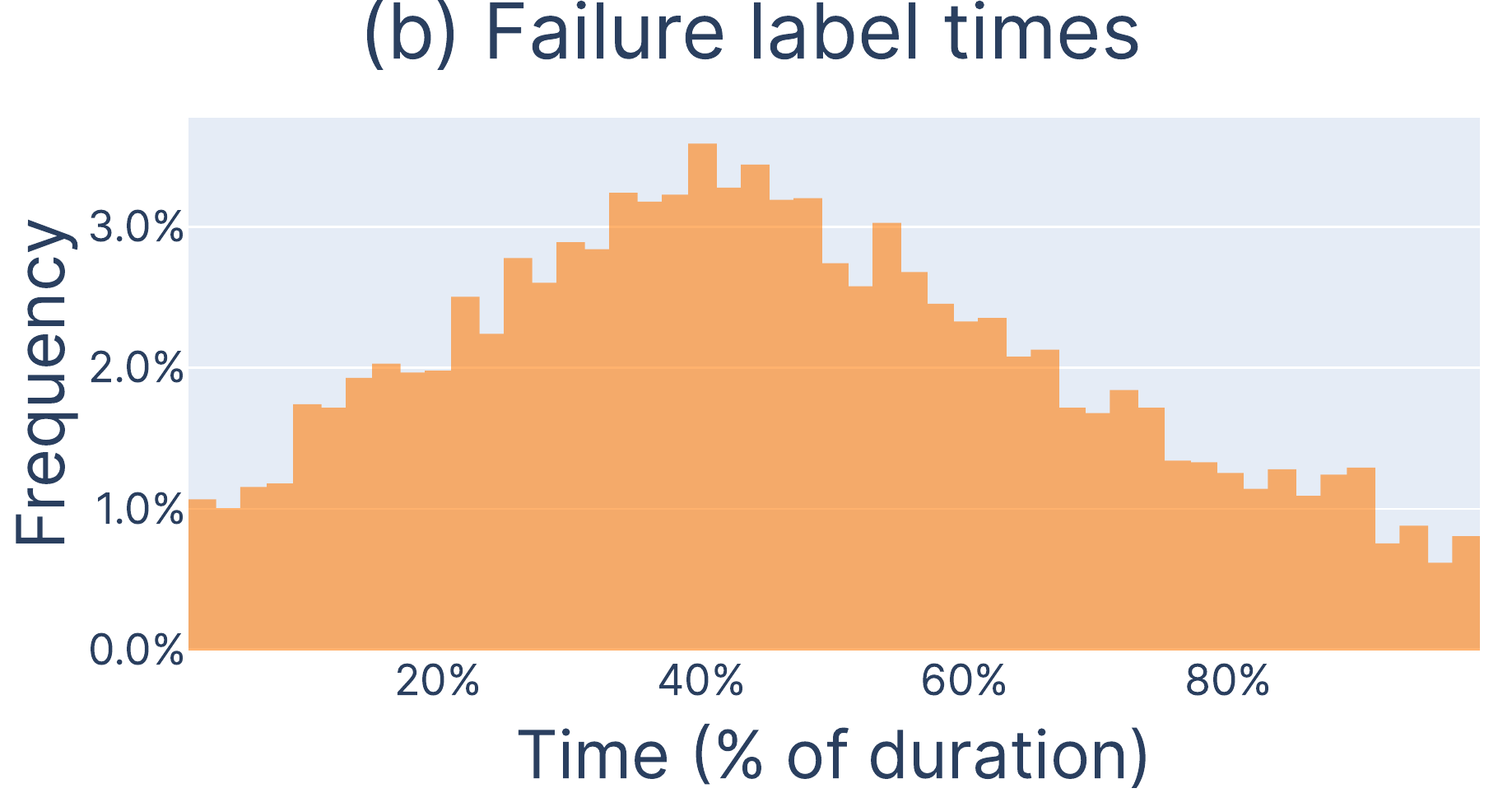}
\phantomsubcaption\label{fig:lbl_rel_dist}
\end{subfigure}
\begin{subfigure}[t]{0.33\linewidth}\vskip 0pt
\includegraphics[width=\linewidth]{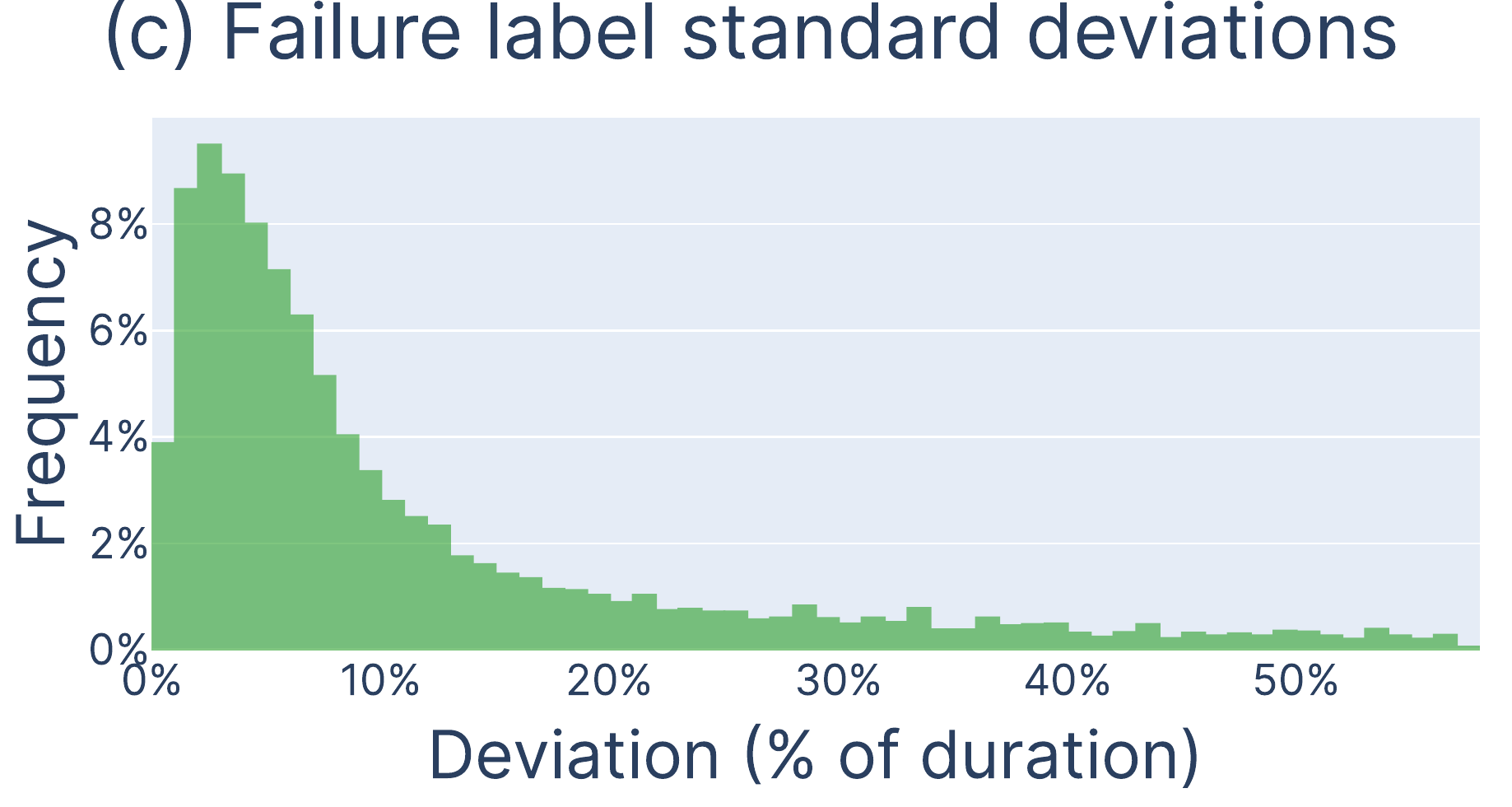}
\phantomsubcaption\label{fig:lbl_rel_stdev}
\end{subfigure}
\begin{subfigure}[t]{0.5\linewidth}\vspace{-1em}
\includegraphics[width=\linewidth]{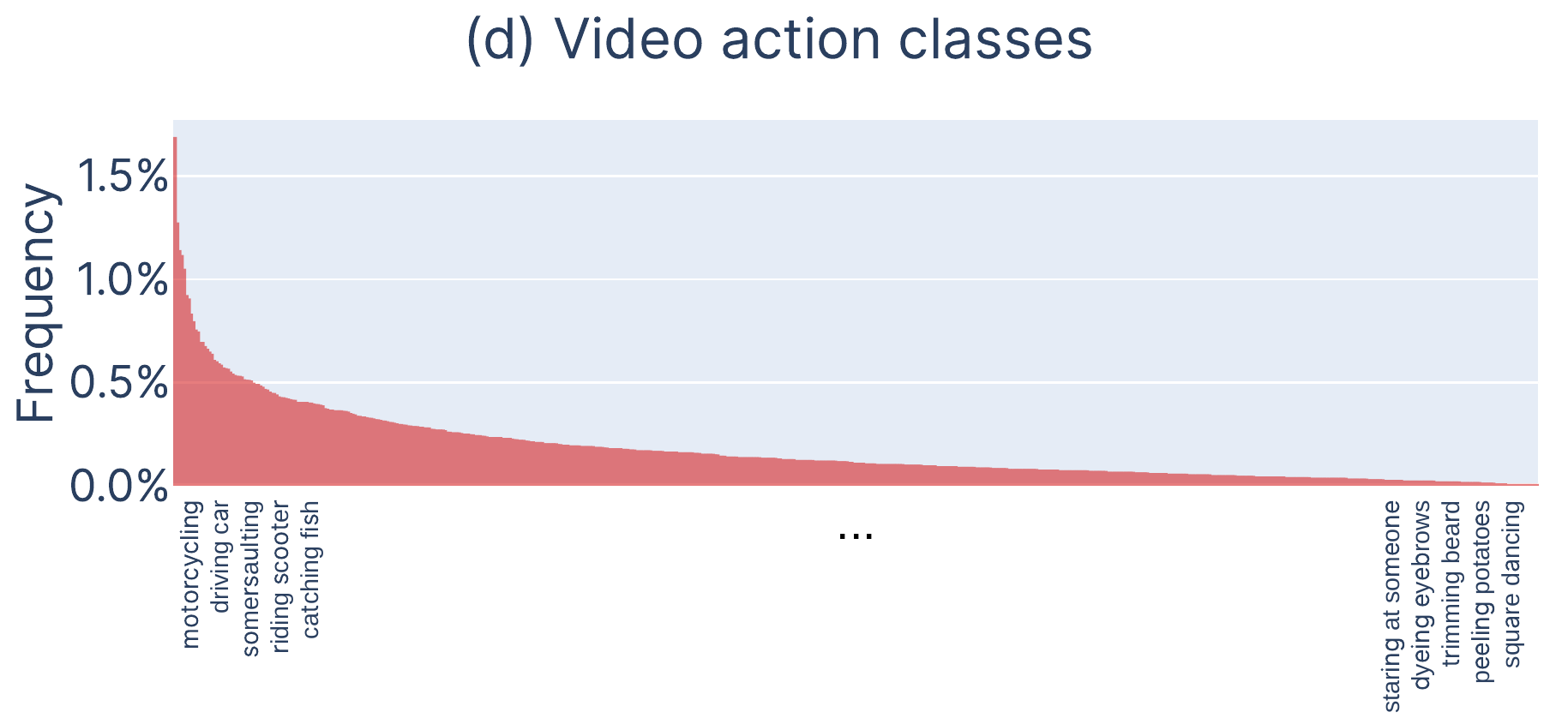}
\phantomsubcaption\label{fig:kinetics_dist}
\end{subfigure}
\begin{subfigure}[t]{0.5\linewidth}\vspace{-1em}
\includegraphics[width=\linewidth]{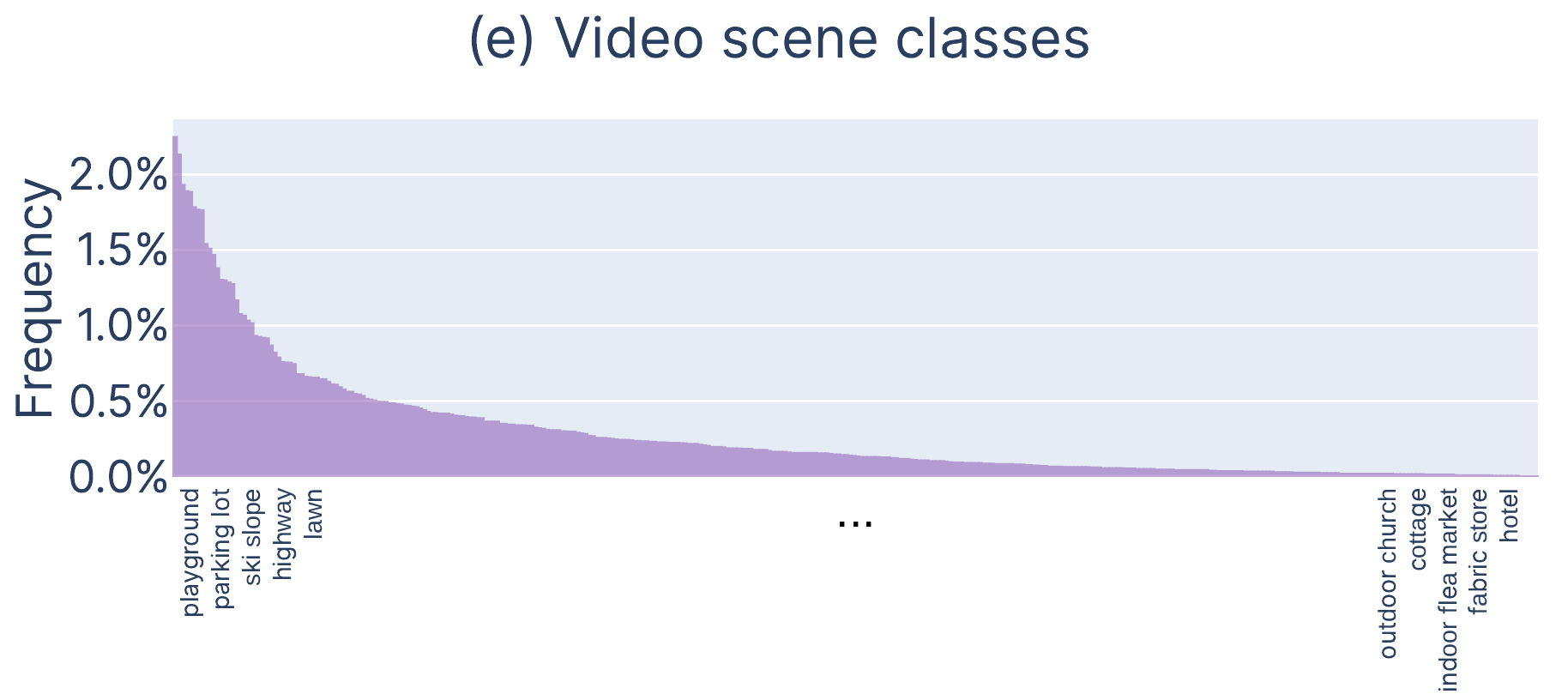}
\phantomsubcaption\label{fig:places_dist}
\end{subfigure}
\vspace{-3em}
\caption{\textbf{Dataset Statistics:} We summarize our dataset with the (\textbf{a}) distribution of clip lengths, (\textbf{b}) the distribution of temporal locations where failure starts, and (\textbf{c}) the standard deviation between human annotators. The median and mean clip lengths are 7.6 and 9.4 seconds respectively. Median standard deviation of the labels given across three workers is 6.6\% of the video duration, about half a second, suggesting high agreement.  We also show the distribution of (\textbf{d}) action categories and (\textbf{e}) scene categories, which naturally has a long tail. For legibility, we only display the top and bottom 5 most common classes for each. 
 Figure best viewed on a computer screen with zoom.\vspace{-1em}}
\label{fig:dataset_stats} 
\end{figure*}

\textbf{Video datasets:}  Computer vision has made significant progress in recognizing human actions through video analysis. Critical to this success are datasets of diverse videos released to facilitate this research \cite{schuldt2004recognizing, blank2005actions, kuehne2011hmdb, soomro2012ucf101, wang2014action, karpathy2014large, caba2015activitynet, abu2016youtube, nguyen2016open, sigurdsson2016hollywood, kay2017kinetics, goyal2017something, fouhey2018lifestyle, damen2018scaling, gu2018ava, monfort2019moments}. Most modern datasets are intended for discriminating between human activities to perform action classification and localization \cite{poppe2010survey, weinland2011survey, aggarwal2011human, cheng2015advances, kang2016review, kang2016review, asadi2017survey, zhang2019comprehensive}. In our paper, we instead focus on analyzing goal-directed human action \cite{velleman1991intention}, and propose a dataset that allows for learning about failed goals and the transition from intentional to unintentional action. Our dataset includes both human errors caused by imperfect action execution (\eg physical interference, limited visibility, or limited knowledge) and human errors due to mistakes in action planning (\eg flawed goals or inadequate reasoning).\par

\textbf{Action recognition and prediction:} Our work builds on a large body of literature in action classification and prediction. Earlier research in action classification \cite{laptev2005space, klaser2008spatio, wang2011action, sadanand2012action, pirsiavash2014parsing} focuses on designing features or descriptors for given input video frames. Recent progress has focused on using deep convolutional networks to solve these tasks, and many methods have been proposed to learn useful feature representations, such as visual information fusion \cite{wang2016temporal, carreira2017quo}, two-stream CNNs \cite{simonyan2014two}, 3D convolutional networks that take in a chunk of video \cite{tran2015learning}, and temporal reasoning for feature extraction \cite{zhou2018temporal, donahue2015long}. In this paper, we base our methods on 3D CNNs.

Previous work which studies future action prediction in video is also relevant to predicting unintentionality \cite{ryoo2011human, pei2011parsing, xie2013inferring, hoai2014max, damen2018scaling}. Many methods rely on action label supervision, along with other auxiliary information, to predict future actions \cite{yu2012predicting, zhao2013online}. Other approaches \cite{vondrick2016anticipating, tran2019back, xu2019self, zolfaghari2019learning} focus on leveraging large unlabeled datasets to learn visual representations useful for action anticipation.\par



\textbf{Self-supervised learning:} Our work uses unlabeled video to learn useful representations without manual supervision. In recent years, self-supervision, which predicts information naturally present in data by manipulating or withholding part of the input, has become a popular paradigm for unsupervised learning. Various types of self-supervised signals have been used to learn strong visual representations, such as spatial arrangement \cite{noroozi2016unsupervised}, contextual information \cite{doersch2015unsupervised, wang2015unsupervised}, color \cite{larsson2017colorization, vondrick2018tracking}, the arrow of time \cite{wei2018learning, xu2019self, hjelm2018learning, lee2017unsupervised, misra2016shuffle, fernando2017self},  future prediction \cite{mathieu2015deep, vondrick2016generating, xue2016visual, oord2018representation}, consistency in motion \cite{agrawal2015learning, jayaraman2015learning}, view synthesis \cite{zhou2016view, zhou2017unsupervised}, spatio-temporal coherence \cite{tung2017self, wang2019learning, dwibedi2019temporal, li2019joint, lai2019self}, and predictive coding \cite{oord2018representation, sun2019contrastive}. Learned representations are then used for other downstream tasks such as image classification, object detection, video clip retrieval, and action recognition. We introduce a new self-supervised pretext task to estimate video speed, which is effective for learning video representations.

\section{The \oops Dataset}\label{sec:dataset}

We present the \oops dataset for studying unintentional human action. The dataset consists of 20,338 videos from YouTube fail compilation videos, adding up to over 50 hours of data. These clips, filmed by amateur videographers in the real world, are diverse in action, environment, and intention. Our dataset includes many causes for failure and unintentional action, including physical and social errors, errors in planning and execution, limited agent skill, knowledge, or perceptual ability, and environmental factors. We plan to release the dataset, along with pre-computed optical flow, pose, and annotations, in the near future. We believe that this dataset will facilitate the development and evaluation of models that analyze human intentionality.

\subsection{Data Collection and Processing}\label{sec:dataset_proc}

We build our dataset from online channels that collate ``fail" videos uploaded by many different users, since the videos they share display unconstrained and diverse situations. Figure \ref{fig:dataset} shows several example frames.

We preprocess the videos to remove editorial visual artifacts. For example, after downloading the long compilation videos from these channels, we must delineate scene boundaries to separate between unrelated clips. We experiment with various such methods and found that \texttt{scikit-video} gives good results.\footnote{We use the scenedet function with method=`edges' and  parameter1=0.7 from \url{https://github.com/scikit-video/scikit-video}} We discard all scenes under 3 seconds long, since they are unlikely to contain a complete scene, as well as all scenes over 30 seconds, since they are likely to contain multiple scenes (due to false negatives in scene detection). Some videos were filmed in portrait orientation but collated in landscape, resulting in a ``letterbox'' effect. We run a Hough line transform to detect these borders, and crop out the border artifacts.

\subsection{Annotation}\label{sec:dataset_anns}

We labeled the temporal locations of failure in the entire test set and some of the training set using Amazon Mechanical Turk \cite{sorokin2008utility}. We ask workers, whom we restrict to a $\geq$99\% approval rating with at least 10,000 approvals, to mark videos at the moment when failure starts to happen (\ie when actions start to become unintentional).

\textbf{Quality Control:} We also use a variety of techniques to ensure high-quality annotation. We repeat annotation three times to verify label quality. We also ask workers to annotate whether the video contained unintentional action or not. We remove all videos where most workers indicate there is no failure or where the failure occurs at the very beginning or end of a video clip (indicating an error in scene detection). The  majority of videos we label pass these checks. To control quality, we also manually label ground truth on a small set of videos, which we use to detect and remove poor annotations.

\textbf{Human Agreement:} We annotated the test set a fourth time, which we use to analyze human agreement on this task. We found that humans are very consistent across each other at labeling the time of failure. The median standard deviation across workers is about half a second, or $6.6\%$ of the video duration. 


\subsection{Dataset Statistics}\label{sec:dataset_stats}
\figref{fig:vid_len} shows the distribution of video clip lengths and \figref{fig:lbl_rel_dist} shows the distribution of failure time labels in the dataset. \figref{fig:lbl_rel_stdev} plots standard deviation of the three labels from different workers, which is around half a second on average. \figref{fig:kinetics_dist} and \figref{fig:places_dist} show the action and scene class distributions, as predicted by models pre-trained on the Kinetics and Places \cite{zhou2017places} datasets. The dataset covers intentions for a variety of scenes and activities.

\begin{figure}[t]
\centering
\includegraphics[width=0.9\columnwidth]{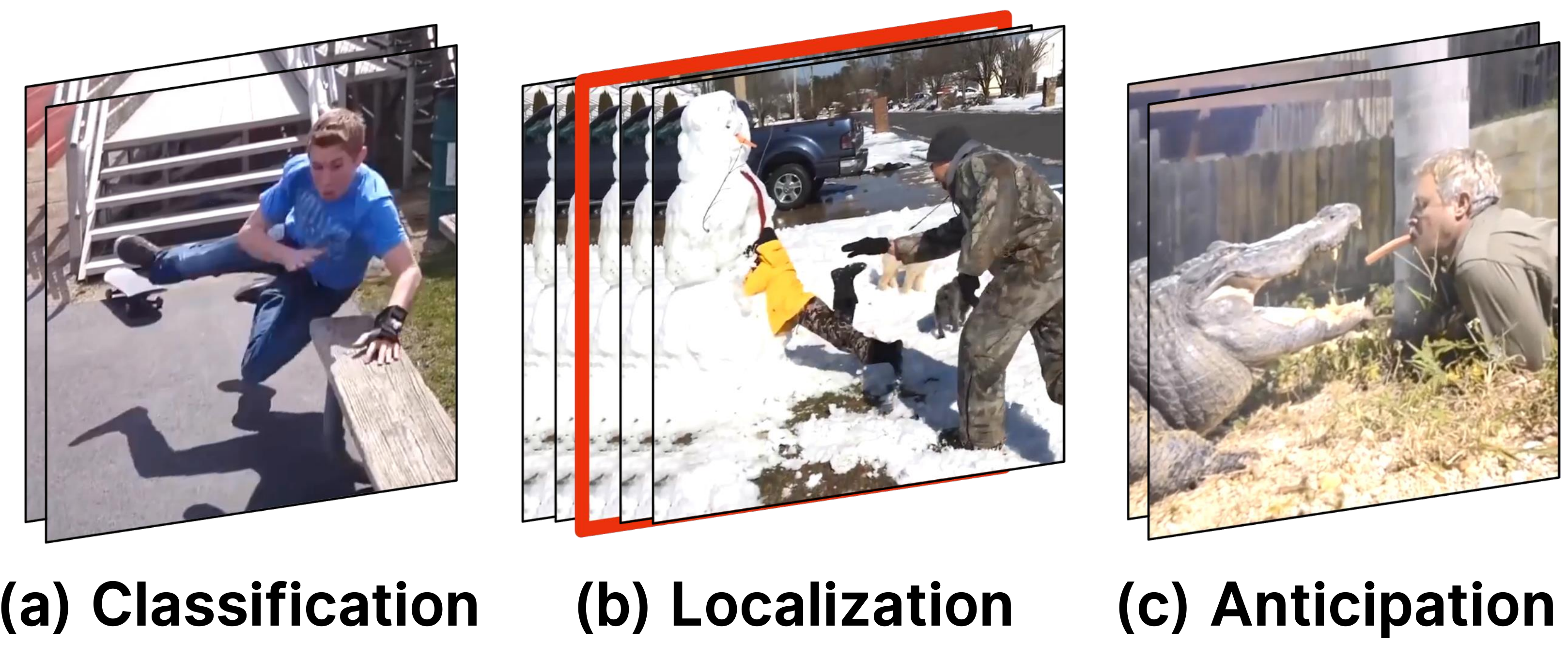}
\vspace{-1em}
\caption{\textbf{Tasks:} Our dataset has three tasks: classification of action as intentional or not,  temporal localization of unintentional action, and  forecasting unintentional action.\vspace{-1em}}
\label{fig:tasks} 
\end{figure}

\subsection{Benchmark}

We use our dataset as a benchmark for recognizing intentional action. We split the dataset into three sets: an unlabeled set of videos for pre-training, a labeled training set, and a labeled test set. The entire dataset contains 20,338 videos, and the labeled training set contains 7,368 videos, which is kept relatively small because the goal of the benchmark is to evaluate self-supervised learning. The test set contains 6,739 videos, which are labeled only for quantitative evaluation. In our benchmark, models are allowed to train on any number of \emph{unlabeled} videos and only a small number of labeled videos. Figure \ref{fig:tasks} shows the tasks for the benchmark.

\section{Intentionality from Perceptual Clues}\label{sec:learning}

We investigate a variety of perceptual clues for learning to predict intentional action with minimal supervision. We can cast this as a self-supervised learning problem. Given incidental supervision from unlabeled video, we aim to learn a representation that can efficiently transfer to different intentionality recognition tasks. 


\subsection{Predicting Video Speed}

The speed of video provides a natural visual clue to learn a video representation.
We propose a self-supervised task where we synthetically alter the speed of a video, and train a convolutional neural network to predict the true frame-rate. Since speed is intrinsic to every unlabeled video, this is a self-supervised pretext task for video representation learning.

\begin{figure}[t]
    \centering
    \includegraphics[width=\linewidth]{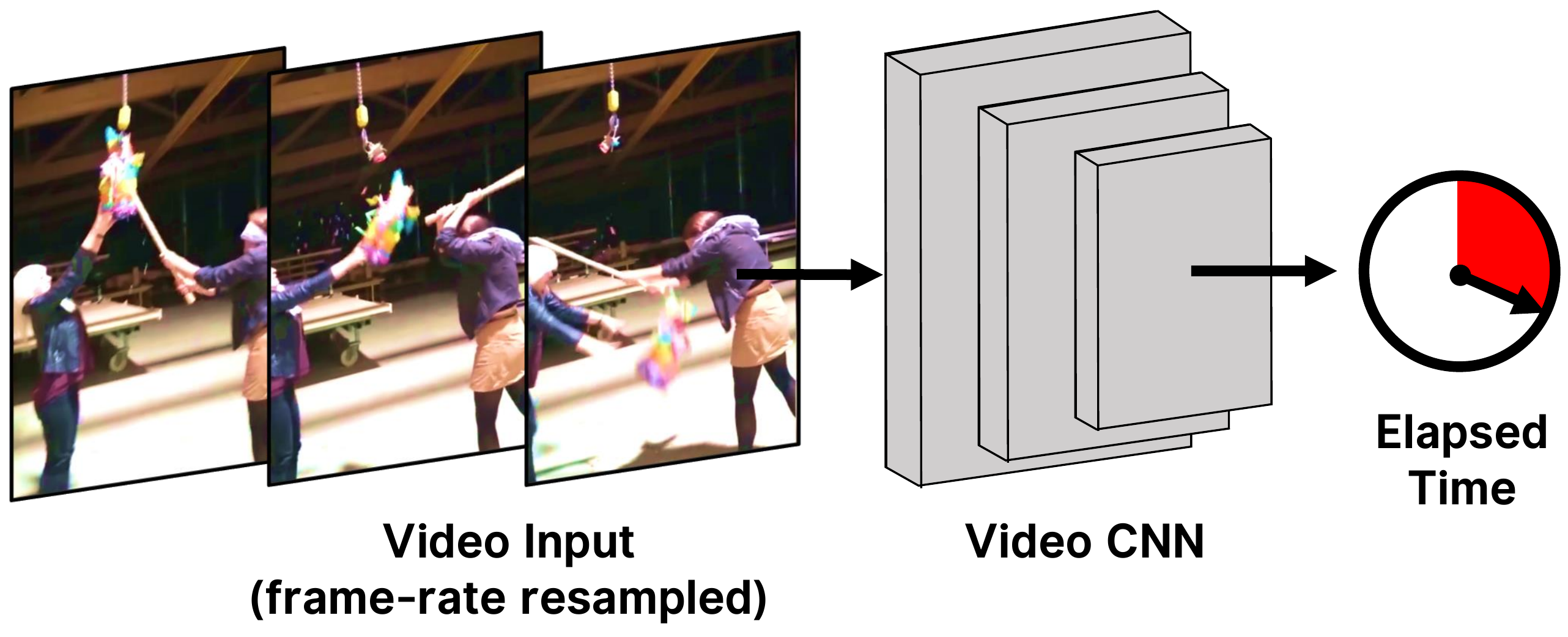}
    \vspace{-2em}
    \caption{\textbf{Video Speed as Incidental Supervision:} We propose a new self-supervised task to predict the speed of video, which is naturally available in all unlabeled video.\vspace{-1em}}
    \label{fig:videospeed}
\end{figure}

Let $x_{i,r} \in \mathbb{R}^{T \times W \times H \times 3}$ be a video clip that consists of $T$ frames and has a frame rate of $r$ frames-per-second.  We use a discrete set of frame rates $r \in \{4,8,16,30\}$ and $T = 16$. Consequently, all videos have the same number of frames, but some videos will span longer time periods than others. We train a model on a large amount of unlabeled video:
\begin{align}
    \min_f \sum_i \mathcal{L}\left(f\left(x_{i,r}\right), r \right)
\end{align}
where $\mathcal{L}$ is the cross-entropy loss function. Figure \ref{fig:videospeed} illustrates this task.

We hypothesize, supported by our experiments, that speed is a useful self-supervisory signal for representation learning. Firstly, estimating the speed requires the model to learn motion features because a single frame is insufficient to distinguish between frame rates. Secondly, this task will require the model to learn features that are correlated to the expected duration of events. For example, the model could detect that a video of a person walking is synthetically sped up or slowed down by comparing it to the average human walking speed. Finally, human judgement of intentionality is substantially affected by video speed \cite{caruso2016slow}. For example, a person leisurely sitting down appears intentional, but a person suddenly falling into a seat appears accidental. Recently, fake news campaigns have manipulated the speed of videos to convincingly forge and alter perception of intent.\footnote{\url{https://www.washingtonpost.com/technology/2019/05/23/faked-pelosi-videos-slowed-make-her-appear-drunk-spread-across-social-media/}}



\subsection{Predicting Video Context}

Since unintentional action is often a deviation from expectation, we explore the predictability of video as another visual clue for intentions. We train a predictive visual model on our unlabeled set of videos and use the representation as a feature space. Let $x_t$ be a video clip centered at time $t$, and both $x_{t-k}$ and $x_{t+k}$ be contextual clips at times $t-k$ and $t+k$ respectively. We learn a predictive model that interpolates the middle representation $\phi_t = f_\theta(x_t)$ from the surrounding contextual frames $x_{t-1}$ and $x_{t+1}$:
\begin{align}
    \max_{f,g} \; \sum_i \log\left(
    \frac{e^{z_t}}
         {e^{z_t} + \sum_{n \in N} e^{z_n} }
    \right)
    \quad \textrm{for} \quad z_j = \frac{\phi^T_j \hat{\phi}_t}{\sqrt{d}}
\end{align}
where $\hat{\phi_t} = g_\theta(\{\phi_{t-k}, \phi_{t+k}\})$ such that $f_\theta$ and $g_\theta$ are convolutional networks. $d$ is the dimension of the representation for normalization, and $N$ is the negative set.

Maximizing this objective corresponds to pulling the features of the target frame, $\phi_t$, closer to the contextual embedding, $\hat{\phi_t}$, while pushing it further away from all other negatives in the mini-batch. This objective is an instance of noise-contrastive estimation \cite{jozefowicz2016exploring} and contrastive predictive coding \cite{oord2018representation,sun2019contrastive,han2019video}, which obtains strong results on other self-supervised learning tasks. We use this as a baseline.

We compute the loss over mini-batches, so the negative set for a given middle clip includes all other clip representations in the mini-batch except itself. We set $g_\theta$ as a two-layer fully-connected network, with hidden dimension 1024, ReLU as activation, and output dimension $d=512$ (same dimension as output of the video encoder $f_\theta$).



\subsection{Predicting Event Order}

We also investigate the order of events as a perceptual clue for recognizing unintentional action. Since unintentional action often manifests as chaotic or irreversible motion, we implement a convolutional model that is tasked with predicting the permutation applied to shuffled input video clips as in \cite{xu2019self,wei2018learning}, which we use as a strong baseline.

We sample 3 clips with a gap of $0.5 \textrm{sec}$ between subsequent clips, so there are $3! = 6$ possible sort orders. We run all clips through a neural network $f_\theta$, which yields a feature vector, then concatenate feature vectors for all pairs of videos and run them through another neural network $g_\theta$, to represent pairwise clip relations. Finally, we concatenate these pairwise representations and input into a third network $h_\theta$ to predict the sort order. The networks $g_\theta$ and $f_\theta$ are both linear layers with a ReLU activation. The output dimensions of $f_\theta$, $g_\theta$, and $h_\theta$ are 512, 256, and 6.

\subsection{Fitting the Classifier}

We use these self-supervised clues to fit a classifier to discriminate action as intentional, unintentional, or transitional. We train the self-supervised models with unlabeled video, and fit a linear classifier with minimal annotation, allowing us to directly compare the quality of the learned representations for recognizing intentionality. 

\textbf{Network Architecture:} We use the same convolutional network architecture throughout all approaches. Since this is a video task, we need to chose a network architecture that can robustly capture motion features. We use the ResNet3D-18 \cite{hara2018can} as the video backbone for all networks, which obtains competitive performance on the Kinetics action recognition dataset \cite{kay2017kinetics}. We input $16$ frames into the model. Except for the video speed model, we sample the videos at 16 fps, so that the model gets one second of temporal context. We train each network for 20 epochs.

\textbf{Classifier:} After learning on our unlabeled set of videos, the self-supervised models will produce a representation that we will use for our intentionality prediction tasks. We input a video into the self-supervised model, extract features at the last convolutional layer, and fit a linear classifier. While there are a variety of ways to transfer self-supervised representations to subsequent tasks, we chose to use linear classifiers because our goal is to evaluate the self-supervised features, following recommended practice in self-supervised learning \cite{kolesnikov2019revisiting}.  We train a regularized multi-class logistic regression using a small amount of labels on the labeled portion of our training set. We formulate the task as a three-way classification task, where the three categories are: a) intentional action, b) unintentional action, and c) transitioning from intentional to unintentional. We define an action as transitioning if the video clip overlaps with the point the worker labeled.

\section{Experiments}\label{sec:exp}

The goal of our experiments is to analyze mid-level perceptual clues for recognizing intentionality in realistic video. To do this, we quantitatively evaluate the self-supervised methods on three tasks on our dataset (classification, localization, and anticipation). We also show quantitative ablations and qualitative visualizations to analyze limitations.

\begin{table}[tb]
\resizebox{\columnwidth}{!}{
\begin{tabular}{l|cc|c}
\toprule
 & \multicolumn{2}{c|}{\textbf{Linear Classifier}} & \textbf{Fine-tune} \\
\textbf{Method} & \textbf{All Labels} & \textbf{10\% Labels} & \textbf{All Labels}\\ \midrule
Kinetics Supervision  & 53.6 & 52.0 & 64.0                  \\ \midrule
Video Speed (ours)  & \textbf{53.4}  & \textbf{49.9}  & \textbf{61.6}             \\
Video Context \cite{oord2018representation}  & 50.0 & 47.2 & 60.3          \\
Video Sorting  \cite{xu2019self}    &     49.8 & 46.5 & 60.2         \\\midrule
Scratch    & 48.2 & 46.2 & 59.4              \\ 
Motion Magnitude     & 44.0 & -  & 44.0             \\
Chance & 33.3  &  33.3 & 33.3  \\ \bottomrule
\end{tabular}
}
\caption{\textbf{Classification Accuracy:} We evaluate performance of each self-supervised model versus baselines. We also compare against a model trained with Kinetics supervision to understand the gap between supervision and self-supervision. This results suggests learning to predict video speed is a promising form of video self-supervision.\vspace{-1em}}
\label{tab:cls}
\end{table}

\subsection{Baselines}

Besides the self-supervised methods above, we additionally compare against several other baselines. 

\textbf{Motion Magnitude:} We use simple motion detection as a baseline. To form this baseline, we compute optical flow \cite{ilg2017flownet} over the videos, and quantify the motion magnitude into a histogram. We experimented with several different bin sizes, and found that 100 bins performed the best. We then fit a multi-layer perceptron on the histogram, which is trained on the labeled portion of our training set to predict the three categories.

\begin{figure}[b]
\includegraphics[width=\columnwidth]{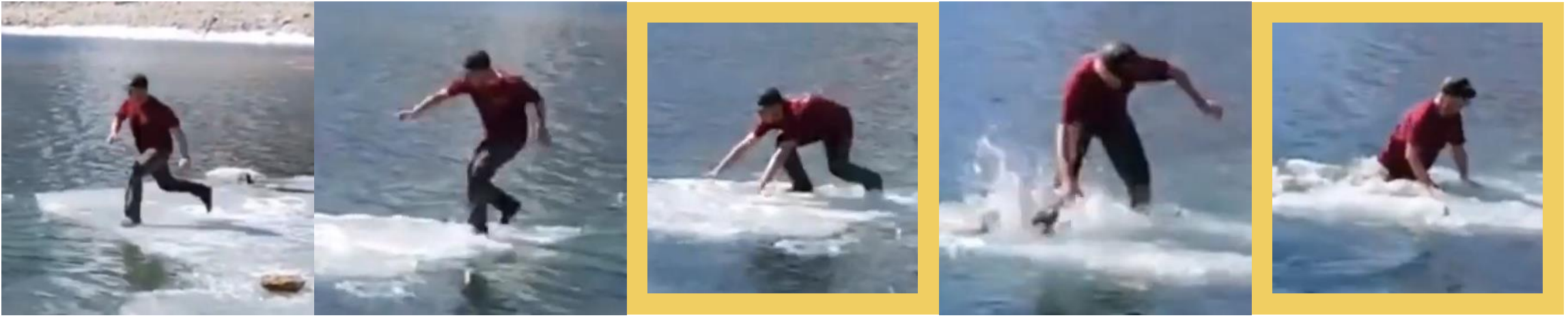}
\vspace{-2em}
\caption{\textbf{Multi-modal Evaluation:} Unintentional actions cascade. For example, in this video, the person ``fails'' twice. To handle this in evaluation, we consider a prediction correct if it is sufficiently near any ground-truth label.}
\label{fig:closest} 
\end{figure}

\textbf{Kinetics Supervision:} We compare against a model that is trained on the full, annotated Kinetics action recognition dataset, which is either fine-tuned with our labeled training set, or used as a feature extractor. Since the model is trained on a large, labeled dataset of over $600,000$ videos, we do not expect our self-supervised models to outperform it. Instead, we use this baseline to understand the gap between supervised and self-supervised methods. 

\textbf{Linear versus Fine-tune:} Unless otherwise noted, the classifier is a linear classifier on the features from the last convolutional layer of the network. However, we also evaluated some models by fine-tuning, which we do to understand the best performance that one could obtain at this task. To fine-tune, we simply use the method as the network initialization, change the last layer to be the three-way classification task, and train the network end-to-end with stochastic gradient descent on our labeled set of videos.

\textbf{Fixed Priors:} We also compare against naive priors. We calculate the mode on the training set, and use this mode as the prediction. Additionally, we use chance. 

\textbf{Human Agreement:} To establish an upper expectation of performance on this dataset, we use a fourth, held-out worker's labels to measure human performance.

\subsection{Classification}\label{sec:exp_clsloc}

We first evaluate each model on a classification task. Given a short video clip, the task is to categorize it into one of the three categories (intentional, unintentional, or transitional). We extract one-second temporal windows in increments of $0.25$ seconds from the testing set.

Table \ref{tab:cls} reports classification accuracy for each method. All of the self-supervised methods outperform baselines, suggesting there are perceptual clues in unlabeled video for intentionality. The model trained with full Kinetics supervision obtains the best performance overall, indicating there is still no substitute for labeled data. However, the gap between the self-supervised models and supervised models is relatively small. For example, the best performing perceptual clue (video speed) is tied with Kinetics when large amounts of labels are available for training a linear layer. We also experimented with the reducing the number of examples in our labeled training set. While accuracy is positively correlated with number of labels, reducing the number of labels by an order of magnitude only causes a minor drop in performance.

\subsection{Localization}

We next evaluate temporal localization, which is challenging because it requires the model to detect the temporal boundary between intentional and unintentional action. We use our classifier in a sliding window fashion over the temporal axis, and evaluate whether the model can detect the point in time that the action switches from intentional to unintentional. The predicted boundary is the one with the most confident score of transition across all sliding windows.  Since videos can contain multiple transitional points, we consider the prediction correct if it sufficiently overlaps any of the ground truth positions in the dataset (Figure \ref{fig:closest}). We use two different thresholds of sufficient overlap: within one second, and within one quarter second.

\begin{figure}[t]
\centering
\includegraphics[width=\linewidth]{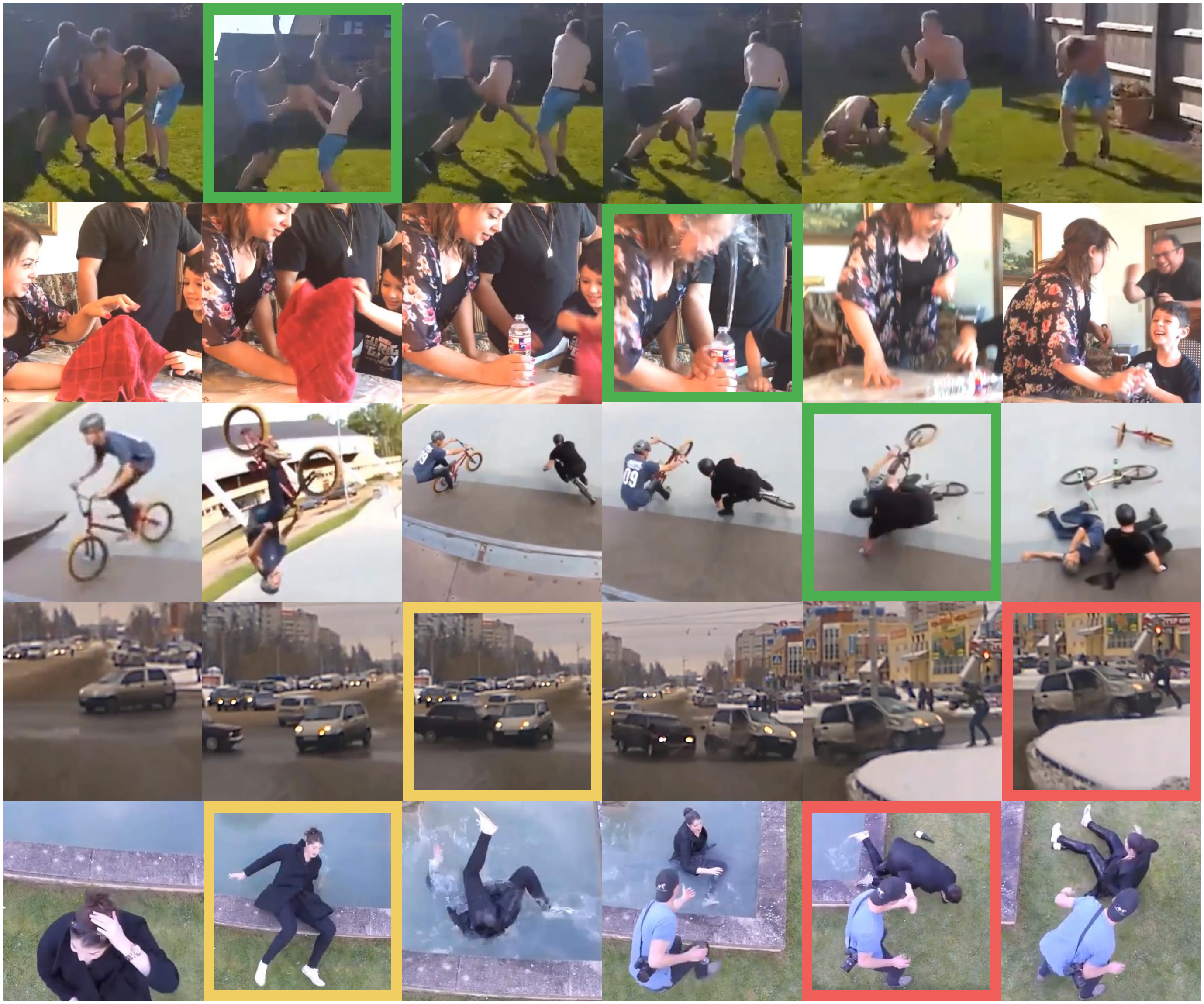}
\vspace{-2em}
\caption{\textbf{Example Localizations:} We show example predictions for localizing the transition to unintentional action. \textcolor[rgb]{0,.56,.12}{Green} indicates a correct prediction (within 0.25 sec). \textcolor[rgb]{.81,0,0}{Red} indicates an incorrect, yet reasonable, prediction. \textcolor[rgb]{.81,.70,0}{Yellow} indicates a missed detection.\vspace{-1em}}
\label{fig:loc_results} 
\end{figure}

Table \ref{tab:loc} reports accuracy at localizing the transition point. For both thresholds, the best performing self-supervised method is video speed, outperforming other self-supervised methods by over 10\%, which suggests that our video speed task learns more fine-grained video features.  Human consistency at this task is high ($88\%$ agreement), however there is still a large gap to both supervised and self-supervised approaches, underscoring the challenge of learning human intent in video. Figure \ref{fig:loc_results} shows a few qualitative results of localization as well as high-scoring false positives. The incorrect predictions our model makes (bottom two rows of Figure \ref{fig:loc_results}) are often reasonable, such as a car hitting a pedestrian on the sidewalk (ground truth: car first hits another car) and person falling when exiting fountain (ground truth: person first falling into fountain).

\begin{table}[tb]
\centering
\begin{tabular}{l | c c}
\toprule
                & \multicolumn{2}{c}{\textbf{Accuracy within}} \\
\textbf{Method} & \textbf{1 sec} & \textbf{0.25 sec} \\ \midrule
Human Consistency   & 88.0   & 62.1    \\ 
Kinetics Supervision (Fine-tune) & 75.9 & 46.7    \\
Kinetics Supervision (Linear)  & 69.2 &  37.8   \\ \midrule
Video Speed (ours) & \textbf{65.3} &  \textbf{36.6}  \\
Video Context \cite{oord2018representation}  &  52.0    & 25.3        \\
Video Sorting \cite{xu2019self}  & 43.3     &    18.3     \\  \midrule
Scratch  & 47.8 &  21.6   \\
Motion Magnitude    &  50.7    & 23.1        \\
Middle Prior       & 53.1 & 21.0    \\
Chance        & 25.9   & 6.8    \\ \bottomrule
\end{tabular}
\vspace{-1em}
\caption{\textbf{Temporal Localization:} We evaluate the model at localizing the onset of unintentional action for two different temporal thresholds of correctness. Although there is high human agreement on this task, there is still a large gap for both supervised and self-supervised models.}
\label{tab:loc}
\vspace{1em}
\centering
\begin{tabular}{@{}l|c}
\toprule
\textbf{Method}           &  \textbf{Accuracy}  \\ \midrule
Kinetics Supervision & 59.7             \\ \midrule
Video Speed (ours)                      & \textbf{56.7}            \\
Video Context \cite{oord2018representation}                     & 51.2                \\
Video Sorting   \cite{xu2019self}           & 51.0            \\  \midrule
Scratch                        & 50.8  \\
Chance                 & 50.0        \\ \bottomrule
\end{tabular}
\vspace{-1em}
\caption{\textbf{Anticipation:} We evaluate performance at predicting the onset of failure before it happens (1.5 seconds into future) by fine-tuning our models. The best performing self-supervised visual clue we considered is video speed.\vspace{-2em}}
\label{tab:ant}
\end{table}

\subsection{Anticipation}\label{sec:exp_ant}

We also evaluate the representation at anticipating the onset of unintentional action. To do this, we train the models with self-supervision as before, but then fine-tune them for a three-way classification task to predict the labels $1.5$ seconds into the future.  Table \ref{tab:ant} reports classification accuracy for the prediction. Features from the the video speed prediction model obtain the best self-supervised performance. However, the model with full Kinetics supervision obtains about 3\% higher performance, suggesting there is still room for self-supervised learning to improve on this task.

\subsection{Analysis}\label{sec:exp_analysis}


Our results so far have suggested that there are perceptual clues in unlabeled video that we can leverage to learn to recognize intentionality. In this subsection, we break down performance to analyze strengths and limitations.

\begin{figure}[tb]
\centering
      \begin{tabular}{crP{1.25cm}P{1.25cm}P{1.25cm}}
 && \multicolumn{3}{c}{\textbf{Predicted Label}} \\ 
\multirow{4}{*}{\rotatebox[origin=c]{90}{\textbf{True Label}}}&& {Intent.}    & {Failure} & {Unintent.} \\ 
 & {Intention}        & \cellcolor{afterfail!62}62.2              & \cellcolor{afterfail!10}10.2       & \cellcolor{afterfail!28}27.6          \\
&{Failure}          & \cellcolor{afterfail!23}22.9              & \cellcolor{afterfail!44}43.9       & \cellcolor{afterfail!33}33.2          \\
&{Unintentional}         & \cellcolor{afterfail!23}23.3              & \cellcolor{afterfail!9}9.3        & \cellcolor{afterfail!67}67.4 \\
& & \multicolumn{3}{c}{(a) Kinetics + Fine-tune} 
\end{tabular}
\\
\vspace{1em}
      \begin{tabular}{crP{1.25cm}P{1.25cm}P{1.25cm}}
\multirow{4}{*}{\rotatebox[origin=c]{90}{\textbf{True Label}}}&& {Intent.}    & {Failure} & {Unintent.} \\ 
 & {Intentional}        & \cellcolor{afterfail!43}43.0 & \cellcolor{afterfail!19.5}19.5 & \cellcolor{afterfail!37}37.4 \\
&{Failure}          & \cellcolor{afterfail!24}24.8 & \cellcolor{afterfail!43}43.5 & \cellcolor{afterfail!31}31.5 \\
&{Unintentional}         & \cellcolor{afterfail!21}21.5 & \cellcolor{afterfail!16}16.0 & \cellcolor{afterfail!62}62.5 \\
& & \multicolumn{3}{c}{(b) Video Speed + Linear} 
\end{tabular}
\vspace{-1em}
\caption{\textbf{Classification Confusion Matrices:} We compare the confusion matrices for (\textbf{a}) Kinetics supervision and (\textbf{b}) self-supervision. One key difference is the self-supervised model often confuses intentional action with the start of failure, suggesting there is substantial room for improving fine-grained localization of intentionality.}
    \label{tbl:conf}
\end{figure}

\textbf{Frequent Confusions:} Figure \ref{tbl:conf} compares the confusion matrices for both the video speed representation and Kinetics supervised representation. In both cases, the most challenging point to predict is the boundary between intentional and unintentional action, which we label the ``Failure'' point. Moreover, a key difference between models is that the self-supervised model more often confuses intentional action with the start of failure. Since the supervised model performs better here, this suggests there is still substantial room to improve self-supervised models at fine-grained localization of intentionality. 

\begin{figure}[t]
\includegraphics[width=\columnwidth]{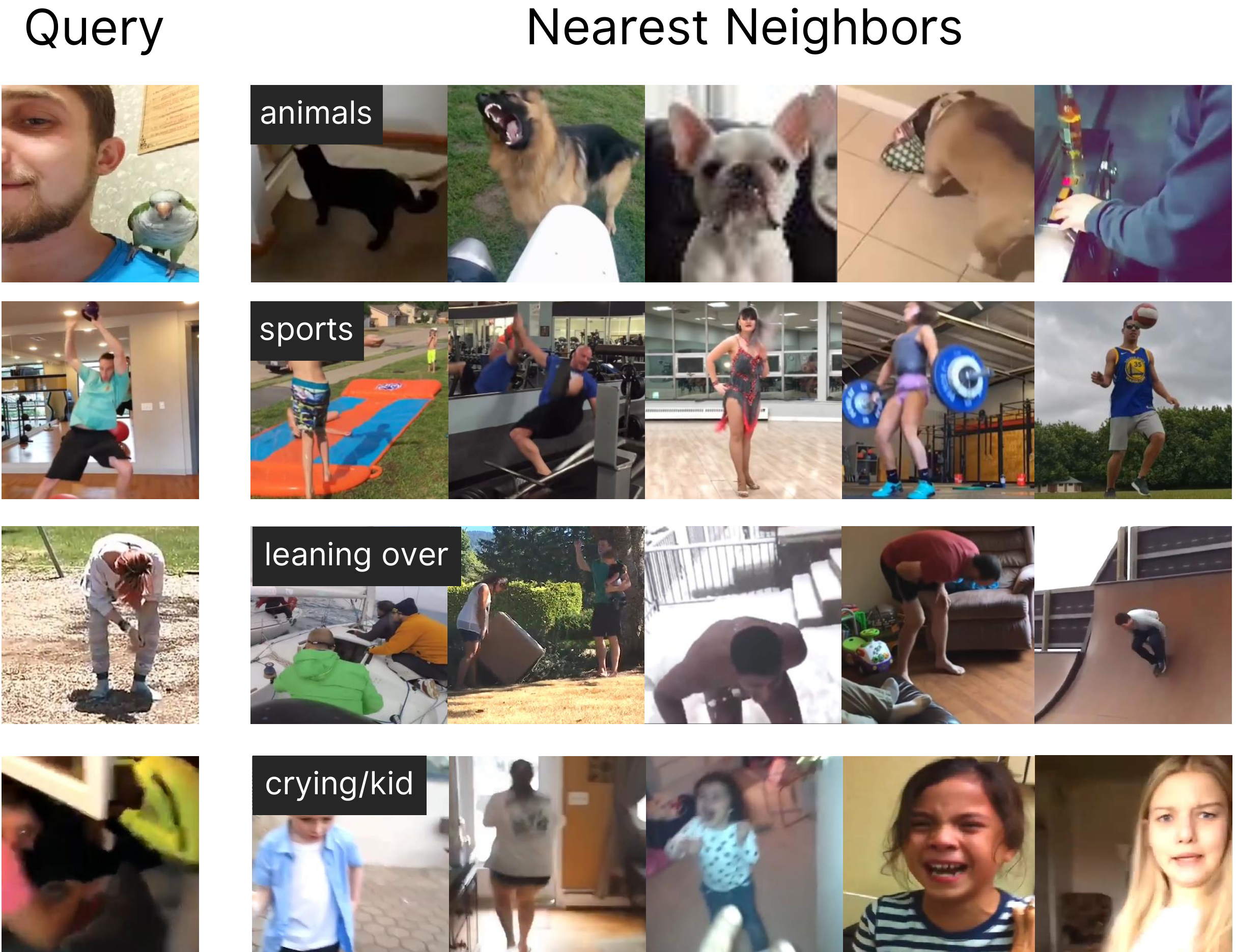}
\caption{\textbf{Nearest Neighbors on Self-supervised Features:} We visualize some of the nearest neighbors from the feature spaced learned by predicting video frame rate. The nearest neighbors tend to be similar activities despite significant variation in appearance.\vspace{-1em}}
\label{fig:nn} 
\end{figure}

\textbf{Visualization of Learned Features:} To qualitatively analyze the learned feature space, Figure \ref{fig:nn} visualizes nearest neighbors between videos using the representation learned by predicting the video speed, which is the best performing self-supervised model. We use one video clip as a query, compute features from the last convolutional layer, and calculate the nearest neighbors using cosine similarity over a large set of videos not seen during training. Although this feature space is learned without ground-truth labels, the nearest neighbors are often similar activities and objects, suggesting that learning to predict the video speed is promising incidental supervision for learning features of activity.

\begin{figure}[t]
\centering
\includegraphics[width=\columnwidth]{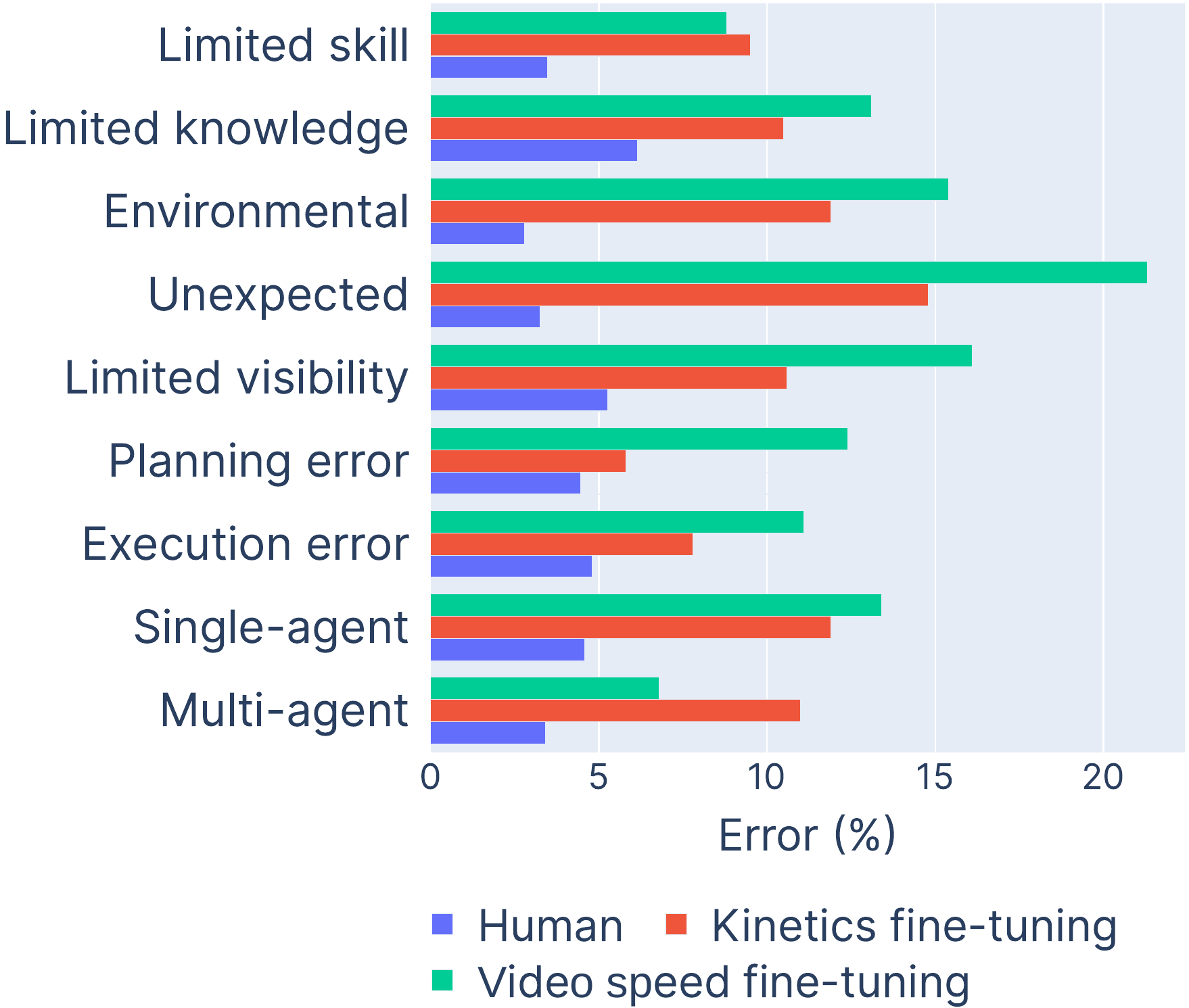}
\vspace{-2em}
\caption{\textbf{Performance Breakdown:} We annotate a diagnostic set of videos into different categories of unintentional action in order to break down model performance and limitations. See text for discussion.\vspace{-1em}}
\label{fig:breakdown} 
\end{figure}

\textbf{Performance Breakdown:}
We manually labeled a diagnostic set of 270 videos into nine types that characterize the cause of unintentional action.
Figure \ref{fig:breakdown} reports performance of models broken down by video type. The most challenging cases for our models are when the unintentional action is caused by environmental factors (such as slipping on ice) or unexpected interventions (such as a bird swooping in suddenly). Moreover, performance is comparatively low when the person in the video has limited visibility, such as due to occlusions, which motivates further work in gaze estimation \cite{recasens2015they,kellnhofer2019gaze360}, especially in video. Another challenge is due to limited knowledge, such as understanding that fire is hot. In contrast, the model has better performance at recognizing unintentional action in multi-agent scenes, likely because multi-agent interaction is more visually apparent. 





\section{Discussion}\label{sec:discussion}

This paper investigates mid-level perceptual clues to recognize unintentional action in video. We present an ``in-the-wild'' video dataset of intentional and unintentional action, and we also leverage the speed of video for representation learning with minimal annotation, which is a natural signal available in every unlabeled video. However, since a significant gain remains to match human agreement, learning human intentions in video remains a fundamental challenge.

\textbf{Acknowledgements:} We thank D\'idac Sur\'is, Parita Pooj, Hod Lipson, and  Andrew McCallum for helpful discussion. Funding was provided by DARPA MCS, NSF NRI 1925157, and an Amazon Research Gift. We thank NVidia for donating GPUs. 

{\small
\bibliographystyle{ieee_fullname}
\bibliography{paper}
}

\end{document}